\documentclass[letterpaper, 10 pt, conference]{ieeeconf}  % Comment this line out if you need a4paper

\IEEEoverridecommandlockouts                              % This command is only needed if 
\usepackage{amsmath}
\usepackage{amssymb}
\usepackage{graphicx} 
\usepackage{xspace}
\usepackage{wrapfig}
\usepackage{algorithm}
\usepackage{algpseudocode}
\usepackage[table,xcdraw]{xcolor}
\title{\LARGE \bf
State and Trajectory Estimation of Tensegrity Robots\\ via Factor Graphs and Chebyshev Polynomials}

\author{%
    Edgar~Granados$^{1}$%
        \thanks{$^{1}$Dept. of Computer Science, Rutgers University, NJ, USA. %
        \{eg585,pm708,ct806,sms840,kb572\}@cs.rutgers.edu. The Rutgers authors are supported in part by NSF under award IIS-1956027. }, %
    Patrick~Meng$^{1}$, % 
    Charles~Tang$^{1}$, %
    Shrimed~Sangani$^{1}$, %
    William~R.~Johnson~III$^{2}$%
        \thanks{$^{2}$Mech. Eng. \& Material Sc. Yale University, New Haven, CT, USA. %
        \{will.johnson,rebecca.kramer\}@yale.edu. The Yale authors are supported in part by NSF under award IIS-1955225. },\\%
    Rebecca~Kramer-Bottiglio$^{2}$, %
    and Kostas~Bekris$^{1}$%
    }%
\newcommand{\cheb}{{Chebyshev}\xspace}

\begin{document}

\maketitle
\thispagestyle{empty}
\pagestyle{empty}

%%%%%%%%%%%%%%%%%%%%%%%%%%%%%%%%%%%%%%%%%%%%%%%%%%%%%%%%%%%%%%%%%%%%%%%%%%%%%%%%
\begin{abstract}
Tensegrity robots offer compliance and adaptability, but their nonlinear, and underconstrained dynamics make state estimation challenging. Reliable continuous-time estimation of all rigid links is crucial for closed-loop control, system identification, and machine learning; however, conventional methods often fall short. This paper proposes a two-stage approach for robust state or trajectory estimation (i.e., filtering or smoothing) of a cable-driven tensegrity robot. For online state estimation, this work introduces a factor-graph-based method, which fuses measurements from an RGB-D camera with on-board cable length sensors. To the best of the authors’ knowledge, this is the first application of factor graphs in this domain. Factor graphs are a natural choice, as they exploit the robot’s structural properties and provide effective sensor fusion solutions capable of handling nonlinearities in practice. Both the Mahalanobis distance-based clustering algorithm, used to handle noise, and the Chebyshev polynomial method, used to estimate the most probable velocities and intermediate states, are shown to perform well on simulated and real-world data, compared to an ICP-based algorithm. Results show that the approach provides high fidelity, continuous-time state and trajectory estimates for complex tensegrity robot motions.
\end{abstract}

%%%%%%%%%%%%%%%%%%%%%%%%%%%%%%%%%%%%%%%%%%%%%%%%%%%%%%%%%%%%%%%%%%%%%%%%%%%%%%%%
% \input{sections/00_progress.tex}
%!TEX root = ../root.tex
\section{INTRODUCTION}

This paper focuses on reliable, continuous-time state estimation for a cable-driven tensegrity robot with limiting assumptions. State estimation remains a crucial component of any robot system, especially for closed-loop operations. There is also a need, however, to provide high-frequency, dynamics-informed state estimation for tasks such as system identification and machine learning. In particular, achieving high-quality simulations of non-conventional robots may require high amounts of reliable labels. While state-of-the-art sensor systems can provide high-frequency data, these systems can be expensive and difficult to operate. Furthermore, sensor fusion algorithms are necessary to make sense of the diverse set of data with varying levels of noise.

\textbf{Motivation} Hybrid soft–rigid robots, such as tensegrities, and other compliant, soft robots, are promising because of their ability to effectively deform to perform desired tasks. Obtaining rich \emph{state} information via state estimation, such as the poses and velocities of each rigid component, can be important for low-level control when it needs to reason about the robot's morphology. Factor graphs have shown promising results for state estimation on classical robotic platforms~\cite{dellaert2017factor}. There has been limited application, however, to soft or tensegrity structures. Factor graphs are appropriate for exploiting the graphical structure of tensegrity robots to obtain sparse representations for real-time applications. They can enable the estimation of dynamical states, thereby improving offline system identification. Additionally, factor graph representations can be used not just for state estimation but also for control~\cite{granados2025kinodynamic}.

\begin{figure}[t]
    \centering
        \includegraphics[width=0.47\textwidth]{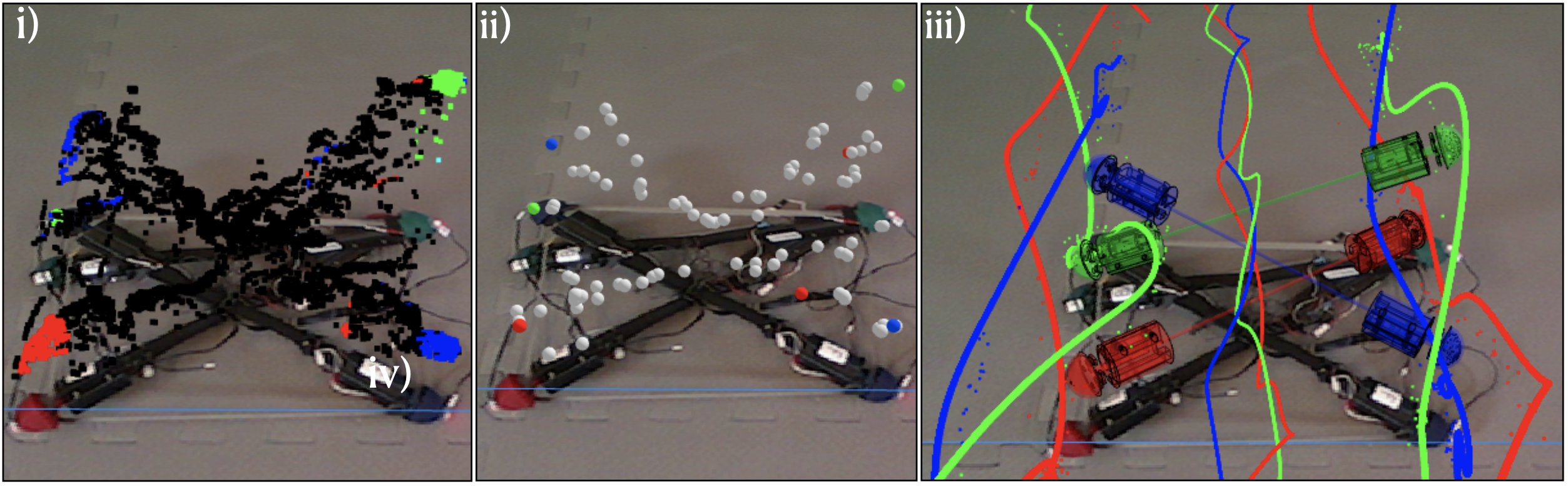}
    \vspace{-.1in}
    \caption{\small Tensegrity estimation: i) Point cloud from RGB-D sensor with self-occlusions. ii) Clustering the points by colors (for visualization, white points show black clusters). iii) Semi-transparent bars show the estimated pose at the example frame, and lines the estimated trajectory via \cheb polynomials.}
    \label{fig:robot_design}
\vspace{-0.3in}
\end{figure}

% Tensegrity robots are more difficult -> soft + rigid 
\textbf{Proposed method and contribution:}
% This paper proposed a factor-graph based estimation algorithm for state and trajectory estimation of a class of tensegrity robots. 
a factor graph-based algorithm for state and trajectory estimation of tensegrity robots. The approach fuses camera observations with on-board cable-length measurements in a unified probabilistic framework that exploits the robot’s structural constraints. A Mahalanobis distance-based clustering algorithm preprocesses noisy and occluded point-cloud data to identify consistent endcap and bar observations, reducing spurious detections before optimization (Fig.~\ref{fig:robot_design}). Within the factor graph, variables represent the six-degree-of-freedom poses of the robot's rigid rods, while factors encode geometric and temporal relationships among components, allowing accurate state estimation even without an explicit actuation model, which can be difficult to define for a tensegrity robot.

An offline approach refines a sequence of discrete state estimates using a Chebyshev polynomial (Fig.~\ref{fig:robot_design}) parameterization obtaining continuous-time trajectories. This enables high-fidelity reconstruction of velocities and intermediate states, essential for system identification and data-driven modeling. The proposed approach represents the first application of factor graphs and Chebyshev polynomials to tensegrity robots, combining robust sensor fusion, statistical clustering, and continuous-time estimation. The choice of \cheb polynomials is motivated by their desirable properties. Similar to splines, \cheb polynomials optimize output in the least-square sense and derivatives are easy to compute. In contrast to splines, it is easy to define \cheb polynomials over Lie-groups~\cite{agrawal2021continuous}.

\textbf{Experiments}
The experiments include a comparison to a previously proposed ICP-based estimation algorithm~\cite{lu20226n}, which depends on manual initialization from a stable tensegrity configuration, limiting its applicability. The methods are evaluated on both a previously presented dataset and a new dataset collected for this effort.
An ablation study uses simulated data from a Mujoco-based tensegrity simulation~\cite{wang2023real2sim2real}. The simulation allows testing the framework with different noise sources and levels to evaluate its robustness. 

% The paper is divided as follows: related work in section~\ref{sec:related_work}, preliminaries and a description of the tensegrity robot in section~\ref{sec:preliminaries}, the factor graph based method along with some comparison points in section~\ref{sec:method}, experiments in section~\ref{sec:experiments}, and conclusions and limitations in sections~\ref{sec:conclusions}.
% This paper introduces i) a factor-graph based estimation algorithm for a type of tensegrity robot, ii) an algorithm to obtain \cheb polynomials representing a trajectory of the robot, iii) an evaluation of the proposed approach in simulated and real data, and iv) comparisons against state of the art algorithms and related approaches.
% \end{itemize}
% Additionally, the framework is tested in the real-world dataset and compered against ground-truth from a mocap system as well as an alternative estimation method~\cite{lu20226n}.
%!TEX root = ../root.tex
\section{RELATED WORK} \label{sec:related_work}

% State estimation
State estimation is crucial in robotics; it is needed online for motion planning and control, while offline estimation is needed for system identification and machine learning. Typically, it has been approached using discrete-time representations, i.e., estimating the state at specific times. These approaches include Bayesian filtering and smoothing. They are popular as they are easy to implement and computationally fast. However, discrete methods are not well-suited under: a) high sensor frequency, and b) need for fine-grained estimation of intermediate states. High sensor updates can be handled by aggregating observations~\cite{forster2016manifold}, interpolating  intermediate states or sacrificing robustness and accuracy.

% Continuous State estimation
Continuous-time state estimation~\cite{talbot2025continuous}  represents the underlying process as a continuous function. This can be advantageous for data-intensive processes, such as system identification and machine learning. 
Popular continuous-time representations include Gaussian Processes and Splines, with recent interest on Chebyshev polynomials \cite{agrawal2021continuous}.
% A urvey~\cite{talbot2025continuous} for more in-depth information.

\textbf{Tensegrity structures} leverage a set of interconnected rods under compression and cables under tension, evenly distributing forces. This property allows tensegrity robots to be  impact resistant, in contrast to traditional  robots \cite{shah2022tensegrity}. Applications include exploration~\cite{vespignani2018design,rieffel2018adaptive}, planetary landers \cite{vespignani2018design}, and deployable structures \cite{meng2021}.
On the other hand, the highly nonlinear, coupled, and underconstrained properties, make state estimation difficult. These characteristics motivate the development of specialized techniques that explicitly account for the unique structure of tensegrity systems.

\textbf{Alternative frameworks} include~\cite{lu20226n}, where first, ICP relates endcap points to a point cloud. Then, endcap and rods are jointly optimized via a cost function accounting for point-cloud error, cable length errors, and ground-plane contact. Proprioceptive estimation via IMU and encoders with geometric structure constraints~\cite{tong2025tensegrity} has also been explored. 
% First, the robot’s shape is obtained by enforcing the known geometric relationships of the tensegrity structure in an optimization framework, yielding endcap positions in body frame. Next, the reconstructed shape and contact information are the input of a contact-aided Invariant Extended Kalman Filter to estimate the global pose in real time. % The estimator achieved an average drift of approximately 4.2\%.

Research has also been done on the use of simulators for tracking tensegrity robots. Wang et al. \cite{wang2023real2sim2real} introduced a differentiable physics-engine tailored to cable-driven tensegrity structures and a Real-to-Sim-to-Real (R2S2R) pipeline that uses real robot trajectory data for system-identification of simulation parameters then generates locomotion policies in simulation and deploys them on hardware. Chen et al. \cite{chen2024learning} proposed a learned simulator based on graph-neural-networks (GNNs) that represented the rods and cables of a tensegrity robot as a graph, and demonstrates improved accuracy and efficiency over prior first-principles differentiable engines for both 3-bar and 6-bar tensegrity systems. 

% % Tensegrity + State estimation 
% \begin{itemize}
%     \item Previous approach, using ICP+ to estimate the state of the rods \cite{lu20226n}
%     \item EKF to estimate the \textit{centroid} of the robot. In contrast to the proposed approach, this one relies on data from IMU, cable lengths and end-cap positions \cite{tong2025tensegrity}.
%     \item Use of estimation for system identification \cite{wang2023real2sim2real, chen2024learning}
% \end{itemize}
% \cite{lilge2024state} -> factor graph + soft robot

% \textbf{Tensegrity Robot and System}
% The tensegrity robot consists of three bars, each with colored endcaps (red, green and blue). Additionally, the robot has nine length sensors connected between endcap pairs which give information about the shape of the robot but not about its location. A real-sense camera located on top of scene gives observations of the state of the robot. Additional cables connect pairs of bars and actuate the robot by changing its length. In this work it is considered that there is no access to an actuation model and therefore, it is not included as part of the state estimation.
% =======

\begin{figure}[t]
    \centering
        \includegraphics[width=0.47\textwidth]{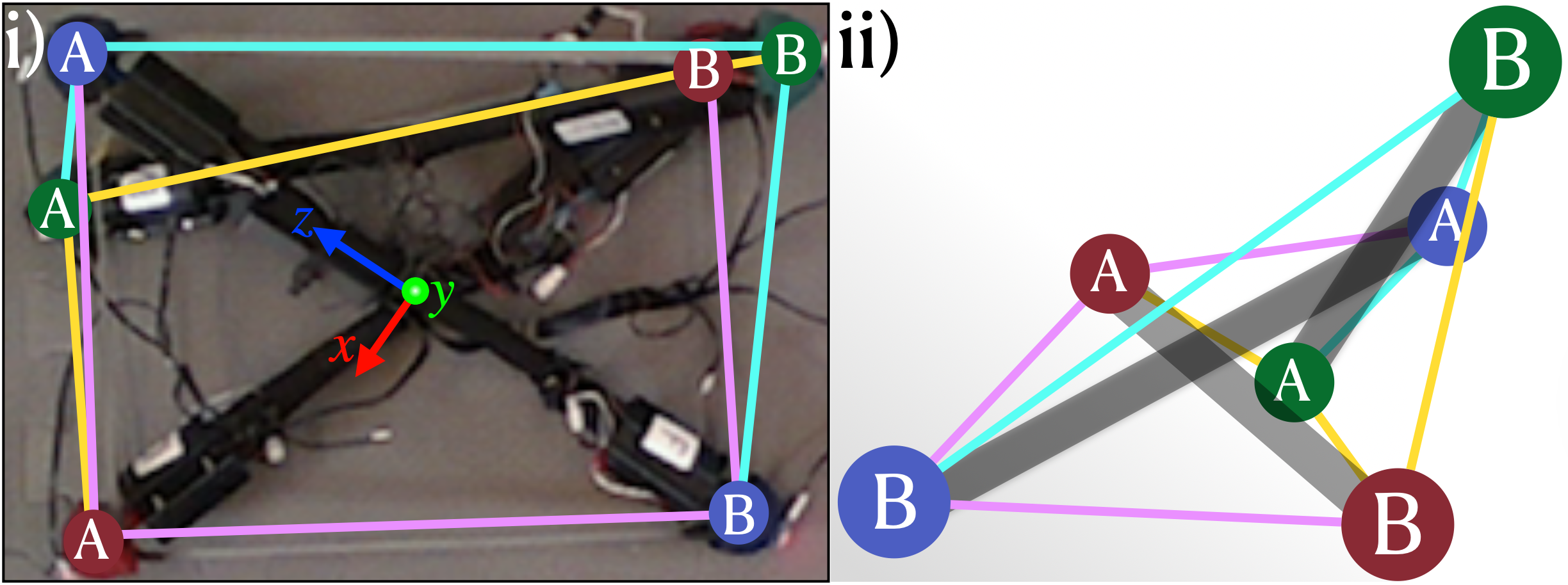}
    \vspace{-.1in}
    \caption{\small The tensegrity robot structure: i) A bar's local frame (the blue bar's frame is shown) is defined on the middle of the bar with endcaps along the $Z$-axis. Each bar has 2 colored endcaps, labeled $A$ or $B$. Endcaps are connected by a total of 9 sensors (6 short and 3 long), shown as color lines. ii) Side view diagram of the robot, where  endcaps $A$ are behind endcaps $B$. Color lines show the \textit{elastic} sensors and black lines show the \textit{rigid} black bars.}
    \label{fig:robot_structure}
\vspace{-0.3in}
\end{figure}

\section{Tensegrity Robot and System}

This work employs a tetherless, mobile tensegrity robot designed to be lightweight and impact-resistant \cite{open_source}. The open-source design is composed of 3d printed and off-the-shelf components. The structure comprises three identical bars differing only in the endcap color. Nine capacitance sensors (cable sensors) span pairs of endcaps (Fig.~\ref{fig:robot_structure}), providing information about the robot’s shape, not its world position. The six short sensors are actively actuated using parallel wires attached to motors, while the other three long sensors are passive elements which elastic properties  restore the robot’s nominal configuration. 
The actuated edges allow the robot to deform, moving along supporting triangles.
% Figure~\ref{fig:robot_design} shows these sensors on the real robot (i) and in a diagram (ii).  
% (the endcaps on the ground that support the robot). 

A RealSense camera positioned above the scene captures RGB-D images, providing observations of the robot’s configuration. A typical control pipeline involves state estimation that fuses RGB-D with on-board sensors. Motion planning and control methods then use the estimation during execution. In conventional tensegrity control, this execution stage depends on an actuation model that maps control inputs to the resulting state. However, this work assumes that no actuation model is available. Consequently, the relationship between actuator commands and structural motion is treated as unknown, and the state estimation focuses exclusively on reconstructing the robot’s shape from sensor and visual data.

% This work considers there is no access to an actuation model and therefore, it is not part of the state estimation.
% >>>>>>> a46a3d84b7960a588440b3bd3dbbfad11ded1af3

%!TEX root = ../root.tex
\section{PRELIMINARIES} \label{sec:preliminaries}

\begin{figure*}[t]
    \centering
		\includegraphics[width=0.95\textwidth]{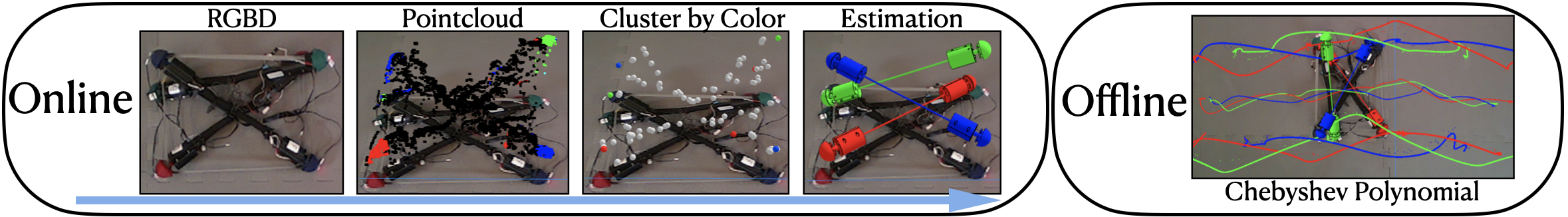}
    \vspace{-.1in}
    \caption{\small Estimation process: from left to right,  images are converted to a point cloud. Points are clustered by color and a factor graph estimates the robot's state.  When a trajectory finishes,  an offline process computes the most likely trajectory using \cheb polynomials. 
    }
    \label{fig:full_process}
\vspace{-0.25in}
\end{figure*}

Consider a robot with state space $\mathbb{X}$, navigating a workspace $\mathbb{W}$ starting at some state $x_0$. The (unknown) {\bf dynamics} govern the robot's motions. An (unknown) controller induces a {\bf trajectory} as a sequence of states $(x_i, x_{i+i},\dots)$.

The robot has access to noisy sensors that provide discrete {\bf measurements}. A measurement from sensor $j$ at step $i$ partially informs the robot's state according to $z^j_i = h^j(x_i) + \epsilon$; where the observation model $h^j(\cdot)$ relates the true state $x_i$ to a measurement with gaussian noise $\epsilon$. Furthermore, it is assumed that the observation model can be analytically obtained or at least approximated. A state estimation process uses measurements at time step $i$ from multiple sensors $Z_i = \{z^j_i, z^{j+1}_i, \dots\} $ to compute an {\bf estimated state} $\bar{x}_i$. Alternatively, a sequence of measurements $(Z_i, Z_{i+1}, \dots)$ can be used to compute an {\bf estimated trajectory}.

{\bf Problem Definition:} Given observations $Z_i$, the first objective is computing accurate, online estimations of the latest state. The secondary objective is to use the sequence of measurements $(Z_i, Z_{i+1}, \dots)$ to compute an estimated trajectory that explains the robot movement. 
% A secondary objective is to obtain a high-fidelity representation of the trajectory that can be query at any point in time. 

Consider $q_i \in \mathcal{M}$ where $\mathcal{M}$ is a \textbf{Lie group} of dimension $m$ and $\dot{q}_i \in \mathbb{R}^m ( \cong T_{q_i}\mathcal{M} )$ is an element in the tangent space of $\mathcal{M}$ at $q$. 
The function $ \texttt{Exp} :  \mathbb{R}^m \rightarrow \mathcal{M} $ maps vector elements to the manifold with its inverse being $ \texttt{Log} : \mathcal{M} \rightarrow \mathbb{R}^m$. The right operators are: $q^{\prime}_i = q_i \oplus \dot{q}_i = q_i * \texttt{Exp}(\dot{q}_i)$ and $\dot{q}_i = q^{\prime}_i \ominus q_i =\texttt{Log}(q_i^{-1} * q^{\prime}_i)$.
The $\texttt{Between}: \mathcal{M} \rightarrow \mathcal{M}$ operation is defined as $\texttt{Between}(q_a,q_b) = q_a^{-1} \circ q_b$ and computes the element that would \textit{move} $q_a$ to $q_b$. Refer to \cite{sola2018micro} for a more in-depth explanation on Lie groups.
% Forward integration in a Lie group is defined as $ q_{i+1} = q_i \circ \texttt{Exp}( \dot{q}_i \Delta t_i)$.

\textbf{Factor Graphs} are a probabilistic framework that has been successful for sensor fusion, state estimation, and localization problems \cite{dellaert2017factor}. In probabilistic inference, the objective is to compute values $\theta$ given events $e$. The posterior density of $\theta$ is computed via Bayes rule: $p(\theta|e) \propto p(\theta)p(e|\theta)$, where $p(\theta)$ is the prior on $\theta$ and $p(e|\theta)$ the likelihood function. The \textit{maximum a posteriori} (MAP) computes the optimal solution: $ \theta^* = \arg\max_\theta p(\theta|e) = \arg\max_\theta p(\theta) l(\theta;e)$
where the likelihood  $l(\theta;e)= p(e|\theta)$ is the probability of events $e$ given  $\theta$. A general likelihood function for non-linear factor graphs is: $l(\theta;e) \propto \exp(0.5|| h(\theta, e)||^2_{\Sigma})$, where $h$ is a \textit{measurement function} with covariance $\Sigma$. 
Formally, a factor graph is a bipartite graph $FG = (\Theta,\mathcal{F},\mathcal{E})$ with factors $f_i \in \mathcal{F}$ and variables $\theta_i \in \Theta$; the posterior is: $ p(\theta|e)\propto \prod_{n=0}^N f_n (\theta_n)$.
The MAP estimate can be reduced to a nonlinear least squares problem and solved with standard solvers. 

\textbf{Chebyshev Polynomials}
This is intended as a brief introduction on the concepts related to \cheb Polynomials \cite{trefethen2019approximation,talbot2025continuous}. \cheb polynomials are analogous to Fourier series and Laurent polynomials, approximating Lipschitz continuous functions on $[-1,1]$ via: $f(x) = \sum_{k=0}^{\infty}a_k T_k(x) $, where the coefficients are given for $k\geq 1 $ by: $ a_k = \frac{2}{\pi}\int_{-1}^{1} \frac{f(x)T_k(x)}{\sqrt{1-x^2}}dx $; for $k=0$ replace $2/\pi $ for $1/\pi$. For a function with known bounds, it is trivial to map it to $[-1,1]$. 
To ensure convergence and avoid numerical issues, the \cheb polynomial of degree N needs to query the function to approximate at the \cheb points, defined by: $\tau_j = \cos(j\pi/N) 0 \leq j \leq N$.
An advantage of \cheb polynomials is the ability to query at any point via $x_i = {\bf X} \cdot {\bf w_i}$, where ${\bf X} $ is a vector of parameters representing the polynomial and ${\bf w_i}$ is the vector of Barycentric weights \cite{berrut2004barycentric}. 

Approximating a function using a \cheb polynomial can be done in two ways. Given direct access to the function, it suffices to query the function at the \cheb points. However, given arbitrary observations, non-linear numerical optimization can use pseudo-spectral parametrization to minimize the function 
$ \sum_{i=1}^{m}\parallel {\bf X} \cdot {\bf w} - z_i \parallel^2_\Sigma$.
% This can be implemented as a factor graph with one variable per polynomial and as many factors as measurements. 

\cheb polynomials can be extended to approximate functions of higher dimensions by computing ${\bf X}$ as an $M \times N $ matrix where $M$ is the dimension of the function and $N$ is the degree of the polynomial or a function evolving in a Lie Group by constructing the polynomial in the Lie Algebra (the tangent space, which is Euclidean). Then, a state at an arbitrary time can be computed as $ x_i = \texttt{Exp}({\bf X} \cdot {\bf w_i}) $.

% \textbf{Other definitions}
The \textbf{Mahalanobis distance}~\cite{mahalanobis1936generalised} measures how distant a point is from the mean of a multivariate normal $N(\mu,\Sigma)$. Given a measurement $z_i$, the Mahalanobis distance is: $D^2(z_i,\mu,\Sigma) = (z_i-\mu)^T \Sigma^{-1} (z_i-\mu)$. The Mahalanobis distance can be seen as the sum of \textit{n} independent standard normal variables, following a chi-squared distribution with degrees of freedom equal to the dimensionality~\cite{manly2024multivariate,etherington2019mahalanobis}.

%!TEX root = ../root.tex
\section{TENSEGRITY ESTIMATION} \label{sec:method}

% \subsection{Factor Graph-based Estimation}
\begin{table*}[t]
\begin{center}
\caption{\small Factors and variables used for online state estimation of the Tensegrity Robot. }
\vspace{-0.05in}
\resizebox{\textwidth}{!}{% Please add the following required packages to your document preamble:
% \usepackage[table,xcdraw]{xcolor}
% Beamer presentation requires \usepackage{colortbl} instead of \usepackage[table,xcdraw]{xcolor}
\begin{tabular}{|c|c|c|c|}
\hline
\rowcolor[HTML]{C0C0C0} 
\textbf{Factor}                & \textbf{Variables}                                                 & \textbf{Definition}                                                                                                       & \textbf{Condition to add to FG}     \\ \hline
${\scriptstyle f^{bar}}$       & ${\scriptstyle X^{R}_i,X^{G}_i,X^{B}_i}$                           & ${ \scriptstyle min( h^{bar}(X^G_i)=\parallel M_{2,3}*(X^{G}_i)^{-1}*z^b_i \parallel,h^{bar}(X^R_i),h^{bar}(X^B_i))}$   & Black point $z^b_i$ observed  \\ \hline
${ \scriptstyle f^{between}}$  & ${ \scriptstyle  X^j_i,X^j_{i-1} }$                                & ${ \scriptstyle  (X^{j}_{i})^{-1} \circ X^j_{i-1}}$                                                                       & Previous estimation available \\ \hline
${ \scriptstyle f^{endcap}_{(\cdot)}}$   & ${ \scriptstyle  X_i,R^{(\cdot)}_i }$                              & ${ \scriptstyle  h^{endcap}(X_i,R^{(\cdot)}_i) - z^e_i =  X_i (R^{(\cdot)}_i  p^{off}) - z^e_i }$                      & Endcap $z^e_i$ observed       \\ \hline
${ \scriptstyle f^{cable}} $   & ${ \scriptstyle X^{j}_i,X^{k}_i,R^{j(\cdot)}_i, R^{k(\cdot)}_i }$  & ${ \scriptstyle  \parallel h^{endcap}(X^{j}_i,R^{j(\cdot)}_i) - h^{endcap}(X^{k}_i,R^{k(\cdot)}_i) \parallel - z^c_i }$   & Cable $z^c_i$ observed        \\ \hline
${ \scriptstyle f^{R0} }$      & ${ \scriptstyle  R^{jA}_i, R^{jB}_i }$                             & ${ \scriptstyle (R^{jA}_i*R^{jB}_i) \ominus R^{off}}$                                                                     & $f^{endcap}_{(\cdot)}$ or $f^{cable}$ is added \\ \hline
${ \scriptstyle f^{R1} }$      & ${ \scriptstyle  R^{jB}_i, R^{jA}_i }$                             & ${ \scriptstyle  (R^{jB}_i*R^{jA}_i) \ominus R^{off}}$                                                                    & $f^{endcap}_{(\cdot)}$ or $f^{cable}$ is added \\ \hline
${ \scriptstyle f^{R2} }$      & ${ \scriptstyle  R^{jA}_i, R^{jB}_i}$                              & ${ \scriptstyle  R^{jA}_i \ominus (R^{jB}_i*R^{off})} $                                                                   & $f^{endcap}_{(\cdot)}$ or $f^{cable}$ is added \\ \hline
\end{tabular}
% $f^{bar}$       & $X^{R}_i,X^{G}_i,X^{B}_i$ & $min( h^{bar}(X^G_i),h^{bar}(X^R_i),h^{bar}(X^B_i));$ where $ h^{bar}(X^{(\cdot)}_i)= \parallel M_{2,3}*(X^{(\cdot)}_i)^{-1}*z_i \parallel$                    &                        \\ \hline
% $f^{bar}$       & $X^{R}_i,X^{G}_i,X^{B}_i$ & $min(\parallel(M_{2,3}*(X^{R}_i)^{-1}*z_i \parallel, \parallel M_{2,3}*(X^{G}_i)^{-1}*z_i\parallel, \parallel M_{2,3}*(X^{B}_i)^{-1}*z_i)\parallel)$                    &                        \\ \hline
                % &                    &                     &                        \\ \hline
                % &                    &                     &                        \\ \hline
% \begin{tabular}{|c|c|c|}
% \hline
% \rowcolor[HTML]{C0C0C0} 
% \textbf{Factor} & \textbf{Definition} & \textbf{When is added} \\ \hline
% $f^{bar}$       &    $min(&\texttt{norm}(M_{2,3}*((X^{j}_i)^{-1}*z_i))), \texttt{norm}(M_{2,3}*((X^{k}_i)^{-1}*z_i))),\texttt{norm}(M_{2,3}*((X^{l}_i)^{-1}*z_i))))$                 &                        \\ \hline
% $f^{between}$   &                     &                        \\ \hline
% $f^{cable}$     &                     &                        \\ \hline
% $f^{endcap}_{(\cdot)}$ &               &                        \\ \hline
%                 &                     &                        \\ \hline
%                 &                     &                        \\ \hline
%                 &                     &                        \\ \hline
% \end{tabular}

% \end{table}/
}
\label{tab:factors}
\end{center}
\vspace{-0.3in}
\end{table*}

The proposed estimation approach for the tensegrity robot is divided in multiple parts as shown in Fig.~\ref{fig:full_process}. Given RGB-D input, it is transformed into a point cloud. However, given the amount of points, a clustering algorithm is used to reduce the amount of data. Then, a factor graph incorporates all available data to estimate the current state of the robot. Finally, an offline process computes the most likely trajectory of the robot leveraging \cheb polynomials. 

\subsection{Observation Models}
\begin{figure}[t]
    \centering
        \includegraphics[width=0.4\textwidth]{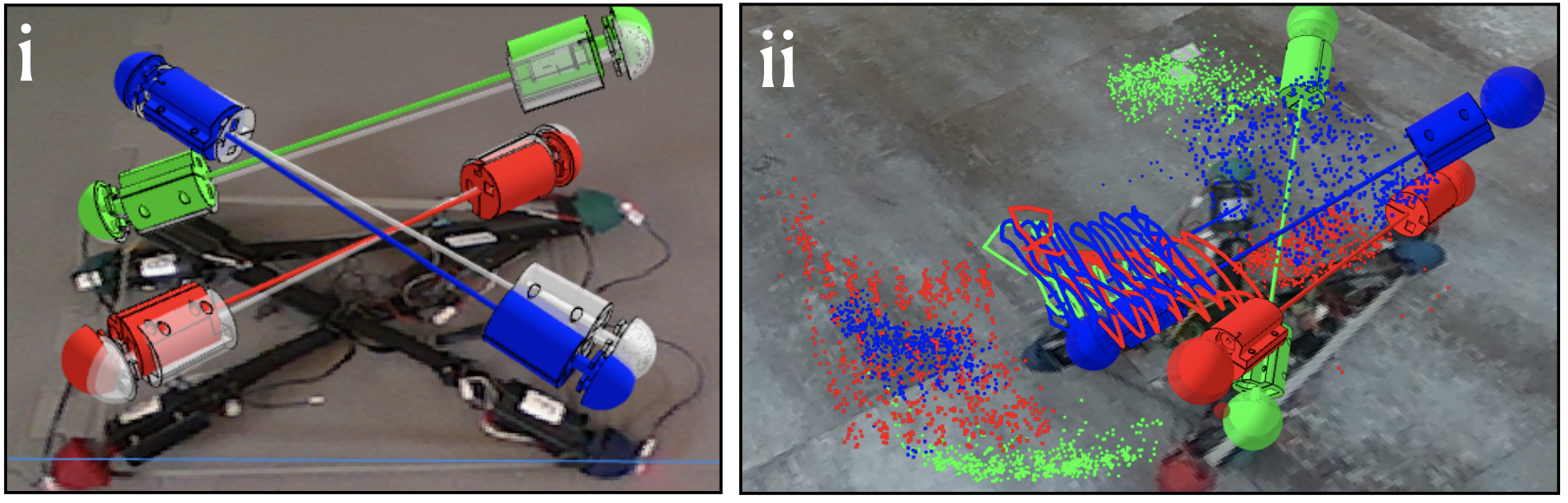}
        \vspace{-.1in}
    \caption{\small i) Estimated state (white) on original dataset, ground-truth (color) poses obtained from mocaps. ii) \cheb polynomial on the new dataset, points show all estimated  endcaps.   }
    \label{fig:pc_cluster_cheb}
\vspace{-0.25in}
\end{figure}

The input are RGB-D images (or point clouds) and cable lengths from the robot. While a point cloud (Fig.~\ref{fig:pc_cluster_cheb}) contains rich information about the scene, it also has varying levels and sources of noise: endcaps may be occluded, similar colors detected elsewhere, and unreliable depth data due to movement of the robot. Furthermore, using raw, high amount of measurements can slow down the optimization process. 

Preprocessing measurements before calling an optimizer is a common approach to handle high frequency observations, specifically IMU data~\cite{forster2015imu}. While camera observations have slower frequency than IMU ones, the amount of data can easily overwhelm an optimizer. For this purpose, this work proposes the use of a Mahalanobis distance-based clustering algorithm to identify both endcaps and bars.

% \begin{wrapfigure}{r}{0.08\textwidth}
% \vspace{-0.1in}
%   \begin{center}
%     \includegraphics[width=0.07\textwidth]{imgs/cluster_fg.png}
%   \end{center}
%   \caption{\small Factor graph solved for clustering.}\label{fig:cluster_fg}
% \vspace{-0.1in}
% \end{wrapfigure}

\begin{figure}[t]
    \centering
		\includegraphics[width=0.45\textwidth]{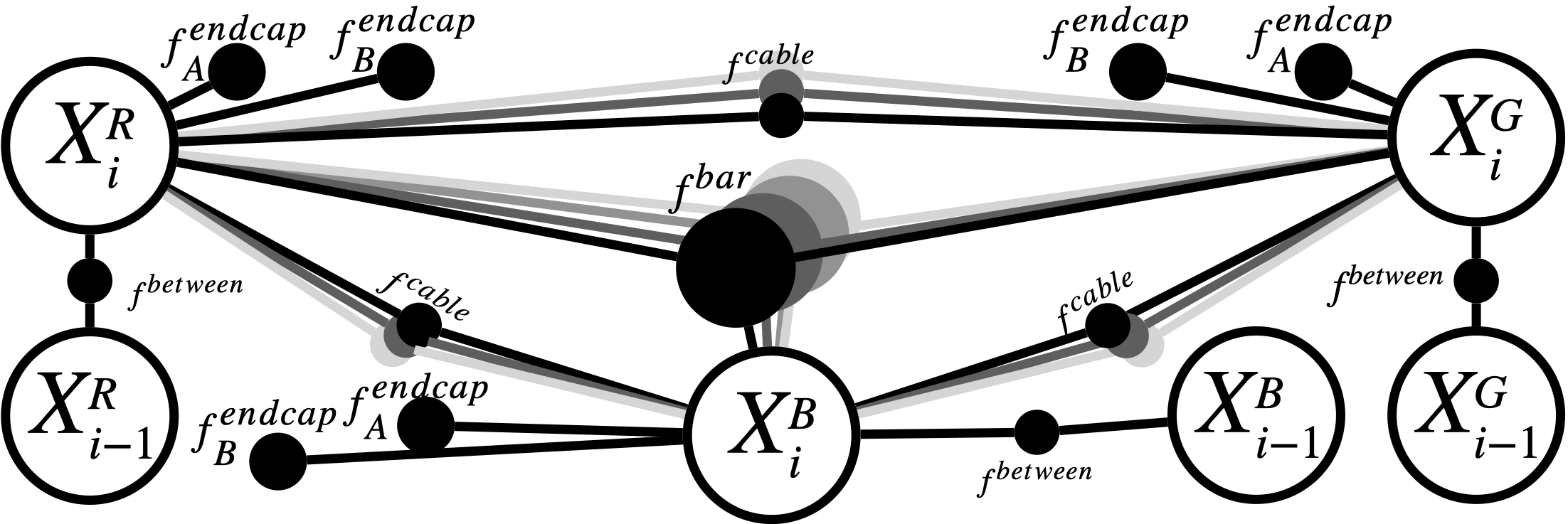}
        \vspace{-0.1in}
    \caption{\small State estimation factor graph: each variable $X_i^k \in SE(3)$ ($k \in \{R,G,B\}$) is a bar pose, and factors (dark discs) relate measurements to variables. Fig. \ref{fig:estimation_fg_rots} expands the subgraph between $f^{cable}$, $f^{endcap}$, and related variables.}
    \label{fig:estimation_fg}
\vspace{-0.3in}
\end{figure}

\textbf{Incremental Clustering} Given a point cloud, the objective is to obtain K clusters best explaining the data. Assume observations ${z_0, z_1, \dots}$, which are samples drawn from a normal distribution $X_k \sim \mathcal{N}(\mu_k, \Sigma_k)$ with unknown mean $\mu_k$, and $\Sigma_k$ determined experimentally. Given a prior estimate of $X_k$ and measurement $z_i$, the problem can be seen in Bayesian framework as finding the probability of a prior on $X_k$ given $z_i$: $p(X_k, z_i)$. Furthermore, assuming independent observations, the probability can be further decompose as: $p(X_k,z_i) = p(x_k)p(z_i|x_k)$ where $p(x_k)$ is the prior on $X_k$. 
% Figure~\ref{fig:cluster_fg} represents this optimization problem as a factor graph.
Algorithm~\ref{alg:clustering} details the process for finding the clusters. 

Solving $p(X_k, z_i)$ gives a new estimate $X^\prime_{k}$, however, this new estimate may not be a \textit{valid} solution. To evaluate the new estimate, the clustering algorithm makes use of the Mahalanobis distance to compute an error: $e_i = D^2(x_k,X^\prime_{k},\Sigma^\prime)+D^2(z_i,X^\prime_{k},\Sigma^\prime)$, where $\Sigma^\prime$ is the approximated covariance on the new estimate. The error $e_i$ gives a measure of \textit{how much} the belief on $X_K$ \textit{moved} by  measurement $z_i$. Given that the Mahalanobis distance follows a chi-squared distribution, if the statistical test $e_i < \chi^2_{n,\alpha} $ is true, the new estimate is accepted. If rejected, the measurement is added to a rejection set, which is then run through the same algorithm recursively. The output of the algorithm can be run through the same algorithm until there is no changes in the output. 

Given the point cloud obtain from the camera, the points are classified by color (endcaps are red, green, or blue and bars are black) and each set is the input of the clustering algorithm. While this algorithm can be computationally expensive for large amounts of points, a significant advantage is its easiness in parallelization. Each color can be clustered independently and, if the set is too big, it can be further divided in a map-reduce approach.
The output of the algorithm is a set of points, which size depends on the spread of the points as well as the covariance $\Sigma$ used.  

\begin{algorithm}
\caption{Cluster}\label{alg:clustering}
\begin{algorithmic}[1]
\Require $Z(t)$ \Comment{Observations}
% \Ensure $y = x^n$
\State $x_k \gets z_0$ \Comment{Initialize with the first measurement}
\State rejected $\gets \{\}$
\For{$\bar{z}_i,\Sigma_i \in Z(t)$}
	\State $X^\prime_{k} \gets $ Solve $p(X_k,z_i) = p(x_k)p(z_i|x_k)$ %\Comment{Solve FG on Fig.~\ref{fig:cluster_fg}}
	\State $\Sigma^\prime \gets $ Marginal on $X^\prime_{k}$ 
	\State $e_i = D^2(x_k,X^\prime_{k},\Sigma^\prime)+D^2(z_i,X^\prime_{k},\Sigma^\prime)$
  \If{ $e_i < \chi^2_{n,\alpha}$} 
  	\State Update prior $p(x_k)$ with new estimate $X^\prime_{k}$
  \Else
  	\State rejected += $(z_i,\Sigma_i)$
  \EndIf 
\EndFor

\State \textbf{return} $\{x_k$, Cluster(rejected)$\}$ \Comment{Recurse if rejected$\neq \emptyset$}
% is Return the current cluster and cluster the rejected (if any)}
\end{algorithmic}
% \vspace{-0.15in}
\end{algorithm}

\subsection{Factor Graph-based Estimation}

\begin{figure}[t]
\vspace{-0.1in}
    \centering
		\includegraphics[width=0.45\textwidth]{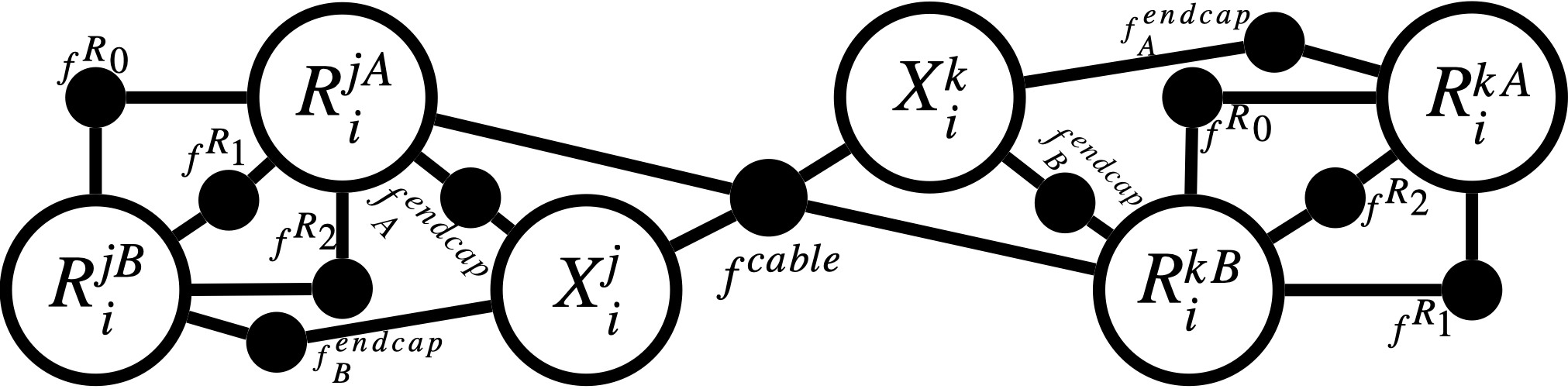}
        \vspace{-0.1in}
    \caption{\small Subgraph 
    relating a cable meassurement between bar endcap $A$ of bar $j$ and endcap $B$ of bar $k$. Factors $f^{R_0},f^{R_1}$, and $f^{R_2}$ are needed for the factor graph to \textit{assign} the unlabeled endcap observations correctly. The full factor graph may contain up to 9 $f^{cable}$ and 6 $f^{endcap}$ factors, with the total variables remaining constant.  }
    \label{fig:estimation_fg_rots}
\vspace{-0.3in}
\end{figure}

The tensegrity state estimation factor graph is shown in Fig.~\ref{fig:estimation_fg} and Fig.~\ref{fig:estimation_fg_rots}, where the corresponding factors are explained in Table~\ref{tab:factors}. Observation factors are added as measurements become available. For instance, a factor $f^{endcap}_B$ is only added if both endcaps of the same type $B$ are detected. At iteration $i$, three variables $X^{(\cdot)}_i \in SE(3)$ represent the  center of each bar. Factors relate observations and priors to  variables. Observations consist of cable lengths (onboard sensing) and bar and endcap points from processed images.

The \textbf{endcap factor} relates an observation of an endcap in world frame $z^e_i \in \mathbb{R}^3 $ to the bar estimate. Given a known, constant offset $p^{off} \in \mathbb{R}^3$, the measurement model endcap \textit{A} is $h^{endcap}_A(X^j_i) = X^j_i  p^{off}$, and for endcap \textit{B} is $h^{endcap}_B(X^j_i) = X^j_i  (-p^{off})$. This model assumes that the endcap has been assigned correctly and is consistent with the other factors. To overcome the correspondence problem, the  model is augmented as: $h^{endcap}(X^j_i,R^{j(\cdot)}_i) = X^j_i  (R^{j(\cdot)}_i p^{off})$
where $R^{j(\cdot)}_i$ is one of $R^{jA}_i$, $R^{jB}_i$ for bar $j$.  The factor $f^{endcap}$ explains the  observation $ z^e_i $, and requires three additional factors ($f^{R_0},f^{R_1}$, $f^{R_2}$) to enforce the symmetry of the bar (Fig.~\ref{fig:estimation_fg_rots}). 
% \vspace{-.05in}
% \begin{align*}
% f^{R0}(R^j_A, R^j_B) &= (R^j_A*R^j_B) \ominus R_{off}\\
% f^{R1}(R^j_B, R^j_A) &= (R^j_B*R^j_A) \ominus R_{off}\\
% f^{R2}(R^j_A, R^j_B) &= R^j_A \ominus (R^j_B*R_{off})
% \vspace{-.05in}
% \end{align*}
% The constant $R^{off}$ explains the rotation (and symmetry) of the two offsets with the bar center.
Factors $f^{R_0}$ and $f^{R_1}$ guarantee that the rotation between $R^{jB}_i$ and $R^{jA}_i$ is $R^{off}$. The  factor $f^{R_2}$ enforces one of $R^{jA}_i$ or $R^{jB}_i$ being $R^{off}$ while the other is the identity.
% To emphasize the other factors, factors R0, R1 and R2 are not depicted in Fig.~\ref{fig:estimation_fg}.
The position \textit{offset} $p^{off}=[0.,0.,0.1625]^T$, and rotation offset $R^{off} = \texttt{quat}(0.,0.,1., 0.)$. That is, each bar measures 1.625cm from center to endcap along the $Z$-axis; $R^{off}$ rotates the bar on its $Y$-axis.

% Chebyshev
\begin{figure*}[t]
    \centering
        \includegraphics[width=0.9\textwidth]{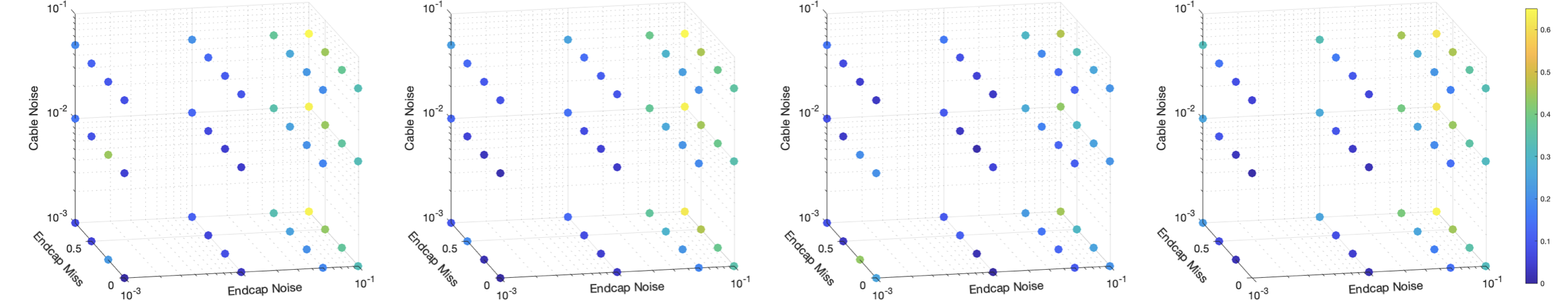}
        \vspace{-0.1in}
    \caption{\small Ablation study of \cheb polynomials on 25 trajectories of trajectories collected from a simulator at 200Hz. Observations are at 30Hz, 20Hz, 10Hz and 5Hz. An endcap or cable observation $z^{(\cdot)}_i$ is generated by adding gaussian noise to the ground-truth value $p^{(\cdot)}_i$ i.e.: $z^{(\cdot)}_i = p^{(\cdot)}_i + N(0,\sigma^{(\cdot)})$. Endcap miss represents the probability of an endcap not being observed, representing occlusions. Each  polynomial is evaluated against the ground-truth state $X^{GT}_i$ at the full resolution of the data, obtaining the error per state as $\texttt{error}_i = | X^{GT}_i  \ominus ({\bf  X \cdot w_i})|$. Colors represent the sum of the elements of the average error vector except for the rotation on the Z-axis. }
    \label{fig:ablation_cheb}
\vspace{-0.25in}
\end{figure*}

The \textbf{cable factor}  $f^{cable}$ relates the measured distance $z^c_i \in \mathbb{R}$ between two endcaps to the corresponding bars' pose.
% The factor $f^{cable}$
% \vspace{-.05in}
% \begin{align*}
% f^{cable}(X^{j}_i,X^{k}_i,R^j_{(\cdot)}, R^k_{(\cdot)}) &= \\ \texttt{norm}(h_{endcap}(X^{j}_i,R^j_{(\cdot)}) &- h_{endcap}(X^{k}_i,R^k_{(\cdot)}) ) - z^c_i
% \vspace{-.05in}
% \end{align*}
% where $\texttt{norm}$ is the euclidean norm. 
At most 9 cable factors (one per sensor) are added per time step; the mapping between sensors and endcaps is known. 

The \textbf{between factor} relates the previous estimate  to the current one.  By computing the relative transform between each bar, the between factor enforces a \textit{similarity} of the estimate $X^{(\cdot)}_{i-1}$ to the new one $X^{(\cdot)}_{i}$. In the absence of an actuation model, these factors penalize big changes between observations. As observations may not have a fix frequency and some observations may be lost, the covariance of this factor is: $\Sigma = z_{dt}I$, where $z_{dt}$ is the time between the previous observation and the current one, resulting in a higher weight on this factor for high frequency updates.

The \textbf{black bars factor} relates clustered bar observation  $z^b_i \in \mathbb{R}^3$ to estimated bar poses, with one factor per observation. 
In contrast to $f^{endcap}$, offset $p^{off}$ is unknown: the observation could be of \textit{any} part of the bar. However, the observation can only be of a single bar, and is enforced by the $f^{bar}$ (see table~\ref{tab:factors}) factor.  
% , where $h^{bar}(X^{(\cdot)}_i)=\parallel M_{2,3}*(X^{(\cdot)}_i)^{-1}*z^b_i \parallel$ 
% \begin{align*}
% f^{bar}(X^{j}_i,X^{k}_i,X^{l}_i) = min(&\texttt{norm}(M_{2,3}*((X^{j}_i)^{-1}*z_i))),\\
% &\texttt{norm}(M_{2,3}*((X^{k}_i)^{-1}*z_i))),\\ &\texttt{norm}(M_{2,3}*((X^{l}_i)^{-1}*z_i))))
% \end{align*}
$f^{bar}$ transforms $z^b_i$ from global to the bar's frame and computes the distance to each bar, using only the minimum one. This factor minimizes the distance to every clustered black point to a single bar.
% and the matrix $M_{2,3}= [I_{2,2};\texttt{Zero}_{2,1}]$ maps a 3D vector into a 2D one. 
% By taking the minimum of the three distances, each of these factor minimizes the distance to every clustered black point to a single bar. 

\textbf{Interface with clustered points} The clustered points may contain invalid or missing observations e.g. colors in the environment similar to the endcaps.
Obtaining multiple observations per bar is not strictly necessary, as the other factors, enforcing the robot's structure, can overcome missing points. However, using multiple conflicting observations is prejudicial to the optimizer. Given a (preprocessed) point cloud, the first step is to compute all valid endcap pairs per color. An endcap pair observation is used if the distance between the two endcaps is within 30\% of the true distance. Single endcap observations are also used. Then, a factor graph as in Fig.~\ref{fig:estimation_fg} is built per combination of valid pairs. All the factor graphs are optimized simultaneously, and the one with the minimum cost is selected. While this step can be computationally expensive, in practice the cluster algorithm rarely returns more than 3 points per color.

\subsection{Chebyshev polynomial Trajectory Estimation}
The online state estimator does not compute velocities , using linear interpolation if intermediate states are needed. Using a \cheb polynomial to estimate the trajectory addresses these problems. The proposed method to estimate the trajectory is as follows: a pseudo-spectral parametrization estimates the trajectory of each endcap, using endcap observations and cable sensors; to avoid having the degree of the polinomial as a parameter, a search is performed to identified the best one; and a polynomial is computed for each bar by querying the endcap polynomials at the \cheb points.

\textbf{Endcap Chebyshev Polynomial} 
To compute a \cheb polynomial explaining the endcap's trajectory, is necessary to adapt both the endcap factor and the cable factor:
\setlength{\abovedisplayskip}{1pt}
\setlength{\belowdisplayskip}{1pt}
\setlength{\abovedisplayshortskip}{1pt}
\setlength{\belowdisplayshortskip}{1pt}
\begin{align*}
f^{endcap\_cheb}(\mathbf{E^{j}}) &= \mathbf{E^j} \cdot \mathbf{w_i} - z^e_i \\
f^{cable\_cheb}(\mathbf{E^{j}},\mathbf{E^{k}}) &= \texttt{norm}(\mathbf{E^j} \cdot \mathbf{w_i} - \mathbf{E^k} \cdot \mathbf{w_i}) ) - z^c_i
\end{align*}

The accuracy of the \cheb polynomial depends on its degree. Given trajectories of variable length and varying  observations, a degree that produces a \textit{good} fit for one trajectory may give bad results for another. To overcome this problem, the observations are divided in two sets, one used for fitting the polynomial and the other for testing. Given an initial degree (0.1 the size of the test set), a line search is done while the test error keeps decreasing. 

\textbf{Bar Chebyshev Polynomial} 
While SE(3) poses could be obtained via the resulting polynomials by using a factor graph, such query is expensive and lacks velocity information. Instead, the bar polynomials is computed by querying the endcap polynomials at the \cheb points via:$f^{bar\_cheb}(\mathbf{X^{j}}) = (\mathbf{X^j} \cdot \mathbf{w_i})  (R^{off}  p ^{off})$
where both $(R^{off}$ and $p^{off}$ are constant.
The degree of the polynomial is computed according to the chopping algorithm in~\cite{aurentz2017chopping}. 
%!TEX root = ../root.tex

\begin{table*}[]
\centering
\caption{\small Experimental results on trajectories for real tensegrity robots (error measured in meter) }
\vspace{-0.15in}
\resizebox{\textwidth}{!}{%
\begin{tabular}{c|ccc|ccc|ccc|}
\cline{2-10}
\multicolumn{1}{l|}{} &
  \multicolumn{3}{c|}{\cellcolor[HTML]{9B9B9B}\textbf{Original Dataset (Long)}} &
  \multicolumn{3}{c|}{\cellcolor[HTML]{9B9B9B}\textbf{Original Dataset (Short)}} &
  \multicolumn{3}{c|}{\cellcolor[HTML]{9B9B9B}\textbf{New Dataset}} \\ \cline{2-10} 
\multicolumn{1}{l|}{} &
  \multicolumn{1}{c|}{\cellcolor[HTML]{C0C0C0}Center of Mass Error} &
  \multicolumn{1}{c|}{\cellcolor[HTML]{C0C0C0}Translation Error} &
  \cellcolor[HTML]{C0C0C0}Rotation Error &
  \multicolumn{1}{c|}{\cellcolor[HTML]{C0C0C0}Center of Mass Error} &
  \multicolumn{1}{c|}{\cellcolor[HTML]{C0C0C0}Translation Error} &
  \cellcolor[HTML]{C0C0C0}Rotation Error &
  \multicolumn{1}{c|}{\cellcolor[HTML]{C0C0C0}Center of Mass Error} &
  \multicolumn{1}{c|}{\cellcolor[HTML]{C0C0C0}Translation Error} &
  \cellcolor[HTML]{C0C0C0}Rotation Error \\ \hline
\multicolumn{1}{|c|}{\cellcolor[HTML]{C0C0C0}ICP} &
  \multicolumn{1}{c|}{0.02837} &
  \multicolumn{1}{c|}{0.03046} &
  0.14599 &
  \multicolumn{1}{c|}{0.00850} &
  \multicolumn{1}{c|}{0.01170} &
  0.10149 &
  \multicolumn{1}{c|}{0.02241} &
  \multicolumn{1}{c|}{0.02885} &
  0.52331 \\ \hline
\multicolumn{1}{|c|}{\cellcolor[HTML]{C0C0C0}Factor Graph} &
  \multicolumn{1}{c|}{0.02965} &
  \multicolumn{1}{c|}{0.03198} &
  0.34468 &
  \multicolumn{1}{c|}{0.02857} &
  \multicolumn{1}{c|}{0.03116} &
  0.37081 &
  \multicolumn{1}{c|}{0.02559} &
  \multicolumn{1}{c|}{0.02484} &
  0.36583 \\ \hline
\multicolumn{1}{|c|}{\cellcolor[HTML]{C0C0C0}Chebychev} &
  \multicolumn{1}{c|}{0.04493} &
  \multicolumn{1}{c|}{0.04805} &
  0.31728 &
  \multicolumn{1}{c|}{0.03419} &
  \multicolumn{1}{c|}{0.03743} &
  0.41592 &
  \multicolumn{1}{c|}{0.03587} &
  \multicolumn{1}{c|}{0.03959} &
  0.41473 \\ \hline
\end{tabular}%
}
\label{tab:experimental_results}
\vspace{-0.2in}
\end{table*}

\section{EXPERIMENTS} \label{sec:experiments}
% First, an ablation study uses a simulated dataset to test the accuracy of the \cheb polynomials in the presence of varying levels of noise. Second, a three different real datasets are used to test the accuracy of the proposed approach.

\subsection{Simulation}
To test the \cheb polynomial estimation, a physics-based simulator of the tensegrity robot is used to collect 25 trajectories of 30 seconds each at 200Hz. For each state $x^{(\cdot)}_i$, the ground-truth endcap position $p^e_i$ and distance between endcaps (cable distance) $p^c_i$ is computed and gaussian noise is added to obtain observations as: $z^{(\cdot)}_i = p^{(\cdot)}_i + N(0,\sigma^{(\cdot)})$. Additionally, a probability of missing the endcap (for example, due to occlusions) is considered. An ablation study (Fig.~\ref{fig:ablation_cheb}) shows the effect of incremental noise in the accuracy of the \cheb polynomial, where each axis represents a different value of an isotropic covariance $\Sigma^{(\cdot)})$. Four different measurement frequencies (30Hz, 20Hz, 10Hz and 5Hz) are tested, however, the evaluation is performed at 200Hz, showing the advantage of the \cheb polynomial.

\subsection{Real Experiments}
Three data sets from a real robot are considered. Datasets \textit{Original Dataset (Long)}, and \textit{Original Dataset (Short)} correspond to data presented in~\cite{lu20226n}, where \emph{Long} includes longer trajectories. A new data set (collected in a different environment, with a different robot) is also tested. The previous datasets contain \textit{mocap} information as ground-truth, while the new data set is manually labeled per second. The old dataset has an average observation frequency of 10 Hz but included gaps between observations of up to one second. The new dataset provides higher temporal resolution, with most sensor streams recorded at 20 Hz.

% Finally, the previous dataset has on average a frequency of 10Hz, with up to a second between observations, where the new data set is collected at the frequency of the camera and sensors -- minimum 20Hz.

The comparison point~\cite{lu20226n} is an ICP-based algorithm with additional constraints for cable sensors. It requires to synchronize observations from different sources (e.g., vision, cable lenghts), which leads to increased noise and missing data without synchronization. It also requires manual initialization of pose estimates from a stable tensegrity configuration.

Table~\ref{tab:experimental_results} presents the ICP-based comparison, the online state estimation, and the \cheb polynomial. Translation error is the distance between estimated positions of each bar and the ground-truth, center of mass is the centroid of the endcaps. Rotation error is $|R^{GT}_i \ominus \hat{R}_i| \in \mathbb{R}^3$, the components of the X and Y axes are summed, as Z is unobservable.% when using only endcap and cable estimations

While the center of mass and translation error of the proposed approach are higher, the rotation error is lower in the new data set. Additionally, the proposed approach does not need manual initialization and can run at a higher frequency. While the \cheb polynomial errors are higher than the factor graph state estimation, the polynomial is a richer representation, better suited for offline system identification and training machine learning algorithms.

% Please add the following required packages to your document preamble:
% \usepackage{graphicx}
% \usepackage[table,xcdraw]{xcolor}
% Beamer presentation requires \usepackage{colortbl} instead of \usepackage[table,xcdraw]{xcolor}

%!TEX root = ../root.tex
\section{CONCLUSIONS}\label{sec:conclusions}
This work presents a factor graph algorithm that reduces information requirements for state estimation of tensegrity robots. The factor graph exploits the system's known structure to achieve fast, accurate localization. Additionally, an offline algorithm computes the \cheb polynomial that best explains the full trajectory. The  rich information provided by the proposed solution can be an advantage for downstream modeling or control solutions and has a less cumbersome initialization relative to a prior approach. Future work can extend this factor graph representation for simultaneous estimation and control~\cite{granados2025kinodynamic}. Furthermore, the proposed factor graph could be extended to different tensegrity structures, such as 6-bar platforms, which utilize similar physical components, i.e., cables and rodes. Additional sensors can be easily incorporated as additional observation factors.

% \printbibliography
\bibliographystyle{IEEEtran}
% \bibliography{refs}
% \bibliographystyle{styles/bibtex/splncs03}

\bibliography{refs.bib}
%%%%%%%%%%%%%%%%%%%%%%%%%%%%%%%%%%%%%%%%%%%%%%%%%%%%%%%%%%%%%%%%%%%%%%%%%%%%%%%%
% \section*{APPENDIX}

%%%%%%%%%%%%%%%%%%%%%%%%%%%%%%%%%%%%%%%%%%%%%%%%%%%%%%%%%%%%%%%%%%%%%%%%%%%%%%%%

% References are important to the reader; therefore, each citation must be complete and correct. If at all possible, references should be commonly available publications.

% \begin{thebibliography}{99}

\end{document}